# Large Language Models to Identify Social Determinants of Health in Electronic Health Records

*Marco Guevara,[1,2] *Shan Chen,[1,2] Spencer Thomas,[1,2,4] Tafadzwa L. Chaunzwa,[1,2] Idalid Franco,[2] Benjamin Kann,[1,2] Shalini Moningi,[2] Jack Qian,[1,2] Madeleine Goldstein, Susan Harper, Hugo JWL Aerts,[1,2,3] Guergana K. Savova,[4] Raymond H. Mak,[1,2] Danielle S. Bitterman,[1,2]

1. Artificial Intelligence in Medicine (AIM) Program, Mass General Brigham, Harvard Medical School, Boston, MA
2. Department of Radiation Oncology, Brigham and Women's Hospital/Dana-Farber Cancer Institute, Boston, MA
3. Radiology and Nuclear Medicine, GROW & CARIM, Maastricht University, The Netherlands
4. Computational Health Informatics Program, Boston Children's Hospital, Harvard Medical School, Boston, MA
* indicates co-first authorship

Corresponding author:
Dr. Danielle S. Bitterman
Department of Radiation Oncology
Dana-Farber Cancer Institute/Brigham and Women's Hospital
75 Francis Street, Boston, MA 02115
Email: Danielle_Bitterman@dfci.harvard.edu
Phone: (857) 215-1489
Fax: (617) 975-0985

Running head: NLP for social determinants of health extraction.
Prior presentations: **None.**

Funding statement: *DSB: Woods Foundation, Jay Harris Junior Faculty Award, Joint Center for Radiation Therapy Foundation. TLC: Radiation Oncology Institute, Conquer Cancer Foundation, Radiological Society of North America. GKS: R01LM013486. RHM: National Institute of Health, ViewRay*
Disclosures: *DSB: Associate Editor of Radiation Oncology, HemOnc.org (no financial compensation, unrelated to this work); Funding from American Association for Cancer Research (unrelated to this work). GKS: none. HA: Advisory and Consulting, unrelated to this work (Onc.AI, Love Health Inc, Sphera, Editas, AZ, and BMS) RHM: Advisory Board (ViewRay, AstraZeneca,), Consulting (Varian Medical Systems, Sio Capital Management), Honorarium (Novartis, Springer Nature).*
Acknowledgments: *The authors thank the Woods Foundation, the Jay Harris Junior Faculty Award, and the Joint Center for Radiation Therapy Foundation for their generous support of this work. The authors also acknowledge financial support from NIH (HA: NIH-USA U24CA194354, NIH-USA U01CA190234, NIH-USA U01CA209414, and NIH-USA R35CA22052), and the European Union - European Research Council (HA: 866504)*

# ABSTRACT

*Objective/Purpose:*
Social determinants of health (SDoH) have an important impact on patient outcomes but are incompletely collected from the electronic health records (EHR). This study researched the ability of large language models to extract SDoH from free text in EHRs, where they are most commonly documented, and explored the role of synthetic clinical text for improving the extraction of these scarcely documented, yet extremely valuable, clinical data.

*Materials/Methods:*
A corpus of 800 patient notes was manually annotated for clinically impactful SDoH categories that are not reliably documented in the EHR as structured data: employment, housing, transportation, parental status, relationship, and social support. Two additional datasets were annotated to assess generalizability: patients' notes from a different clinic in the same hospital system, and MIMIC-III notes. We defined two sentence-level multilabel classification tasks: (1) classifying any SDoH and (2) classifying adverse SDoH. Several fine-tuned transformer-based models including BERT and Flan-T5, were evaluated. We experimented with synthetic data generation to augment the training set, as well as to explore the zero- and few-shot performance of ChatGPT-family models. We assessed for algorithmic bias, and compared SDoH information captured by our models versus structured EHR data.

*Results:*
Our best-performing models were fine-tuned Flan-T5 XL (macro-F1 0.71) for any SDoH, and Flan-T5 XXL (macro-F1 0.70). The benefit of augmenting fine-tuning with synthetic data varied across model architecture and size, with smaller Flan-T5 models (base and large) showing the greatest improvements in performance (delta F1 +0.12 to +0.23). Model performance was similar on the in-hospital system dataset but worse on the MIMIC-III dataset. Our best-performing fine-tuned models outperformed zero- and few-shot performance of ChatGPT-family models for both tasks. These fine-tuned models were less likely than ChatGPT to change their prediction when race/ethnicity and gender descriptors were added to the text, suggesting less algorithmic bias ($p<0.05$). At the patient-level, our models identified 93.8% of patients with adverse SDoH, while ICD-10 codes captured 2.0%.

*Conclusion:*
Fine-tuned language models were able to extract SDoH information from clinic notes, and performed better and more robustly than much larger language models in the zero- and few-shot settings. Compared to structured data, these language models could improve real-world evidence on SDoH and assist in identifying patients who could benefit from resource and social work support.

**INTRODUCTION**

Health disparities have been extensively documented across medical specialties.[1–3] However, our ability to address these disparities remains limited by an insufficient understanding of their contributing factors. Social determinants of health (SDoH), are defined by the World Health Organization as "the conditions in which people are born, grow, live, work, and age [...] shaped by the distribution of money, power, and resources at global, national, and local levels".[4] SDoH may be adverse or protective, impacting health outcomes at multiple levels as they likely play a major role in disparities by determining access to and quality of medical care. For example, a patient cannot benefit from an effective treatment if they don't have transportation to make it to the clinic. There is also emerging evidence that exposure to adverse SDoH may directly affect physical and mental health via inflammatory and neuro-endocrine changes.[5–8] In fact, SDoH are estimated to account for 80-90% of modifiable factors impacting health outcomes.[9]

SDoH are rarely documented comprehensively in structured data in the electronic health records (EHRs),[10–12] creating an obstacle to research and clinical care. Instead, issues related to SDoH are most frequently described in the free text of clinic notes, which creates a bottleneck for incorporating these critical factors into databases to research the full impact and drivers of SDoH, and for proactively identifying patients who may benefit from additional social work and resource support.

Natural language processing (NLP) could address these challenges by automating the abstraction of these data from clinical texts. Prior studies have demonstrated the feasibility of NLP for extracting a range of SDoH.[13–23] Yet, there remains a need to optimize performance for the high-stakes medical domain and to evaluate state-of-the-art language models (LMs) for this task. In addition to anticipated performance changes scaling with model size, large LMs may support EHR mining via data augmentation. Across medical domains, data augmentation can boost performance and alleviate domain transfer issues and so is an especially promising approach for the nearly ubiquitous challenge of data scarcity in clinical NLP.[24–26] The advanced capabilities of state-of-the-art large LMs to generate coherent text open new avenues for data augmentation through synthetic text generation. However, the optimal methods for generating and utilizing such data remain uncertain. Large LM-generated synthetic data may also be a means to distill knowledge represented in larger LMs to more computationally accessible smaller LMs.[27] In addition, few studies assess the potential bias of SDoH information extraction methods across patient populations. LMs could contribute to the health inequity epidemic if they perform differently in diverse populations and/or recapitulate societal prejudices.[28] Therefore, understanding bias is critical for future development and deployment decisions.

In this study, we aimed to characterize the optimal methods and the role of synthetic clinical text for SDoH extraction in the large LM era. Specifically, we used LMs for extracting 6 key SDoH: employment status, housing issues, transportation issues, parental status, relationship, and social support. Because SDoH data are sparsely documented, we assessed the value of adding large LM-generated synthetic SDoH data at the fine-tuning stage. Using a synthetic dataset, we evaluated the performance of state-of-the-art large LMs, including GPT3.5 and GPT4, to identify

SDoH in the zero- and few-shot settings. The potential for algorithmic bias to impact LM predictions was explored. Our results could yield real-world evidence on SDoH, assist in identifying patients who could benefit from resource and social work support, and our research also serves to raise awareness on such an undocumented yet crucial topic.

## MATERIALS AND METHODS

### *Data*

Table 1 describes the patient populations of the datasets used in this study. Our primary dataset consisted of a corpus of 800 clinic notes from 770 patients with cancer who received radiotherapy (RT) at the Department of Radiation Oncology at Brigham and Women's Hospital/Dana-Farber Cancer Institute in Boston, Massachusetts from 2015-2022. We also created two validation datasets. First, we collected 200 clinic notes from 170 patients with cancer treated with immunotherapy at Dana-Farber Cancer, and not present in the RT dataset. Second, we collected 200 notes from 183 patients in the MIMIC (Medical Information Mart for Intensive Care)-III database[29,30], which includes data associated with patients admitted to the critical care units at Beth Israel Deaconess Medical Center in Boston, Massachusetts from 2001-2008. This study was approved by the Mass General Brigham institutional review board, and consent was waived as this was deemed exempt human subjects research.

Only notes written by physicians, physician assistants, nurse practitioners, registered nurses, and social workers were included. To maintain a minimum threshold of information, we excluded notes with fewer than 150 tokens across all provider types. This helped ensure that the selected notes contained sufficient textual content. For notes written by all providers save social workers, we excluded notes containing any section longer than 500 tokens to avoid excessively lengthy sections that might have included less relevant or redundant information. For physician, physician assistant, and nurse practitioner notes, we used a customized medSpacy[31,32] sectionizer to include only notes that contained at least one of the following sections: Assessment and Plan, Social History, and History/Subjective. Please refer to **Supplemental Materials, Appendix A**, for more details on note selection for each dataset.

Prior to annotation, all notes were segmented into sentences using the syntok[33] sentence segmenter as well as split on bullet points "•". This method was used for all notes in the radiotherapy, immunotherapy, and MIMIC datasets for sentence-level annotation and subsequent classification.

*Table 1. Patient demographics across datasets*

| Patients | Radiotherapy Dataset | | | | Out of Domain Validation | | | |
|---|---|---|---|---|---|---|---|---|
| | Total (n=770) | Train Set (n=462) | Development Set (n=154) | Test Set (n=154) | Immuno therapy (n=170) | MIMIC-III (n=183) | Synthetic Validated[a] (n=480) | Synthetic Demo[b] (n=419) |
| **Gender** | | | | | | | | |
| Male | 344 (44.7%) | 210 (45.5%) | 70 (45.5%) | 64 (41.6%) | 75 (44.1%) | 101 (55.2%) | N/A | 168 (40.1%) |
| Female | 426 (55.3%) | 252 (54.5%) | 84 (54.5%) | 90 (58.4%) | 95 (55.9%) | 82 (44.8%) | N/A | 177 (42.2%) |
| Not specified | 0 | 0 | 0 | 0 | 0 | 0 | N/A | 74 (17.7%) |
| **Race** | | | | | | | | |
| White | 664 (86.2%) | 396 (85.7%) | 134 (87.0%) | 134 (87.0%) | 137 (80.6%) | 132 (72.1%) | N/A | 113 (26.9%) |
| Asian | 21 (2.7%) | 11 (2.4%) | 6 (3.9%) | 4 (2.6%) | 9 (5.3%) | 5 (2.7%) | N/A | 106 (21.6%) |
| Black | 37 (4.8%) | 24 (5.2%) | 5 (3.2%) | 8 (5.2%) | 11 (6.5%) | 16 (8.7%) | N/A | 84 (25.7%) |
| Two or more | 3 (0.4%) | 2 (0.4%) | 0 | 1 (0.6%) | 0 | 3 (1.6%) | N/A | 0 |
| Others | 25 (3.2%) | 17 (3.7%) | 5 (3.2%) | 3 (1.9%) | 10 (5.9%) | 1 (0.6%) | N/A | 97 (23.2%) |
| Unknown | 20 (2.6%) | 12 (2.6%) | 4 (2.6%) | 4 (2.6%) | 3 (1.8%) | 25 (13.7%) | N/A | 19 (4.5%) |
| **Ethnicity** | | | | | | | | |
| Non-Hispanic | 682 (88.6%) | 420 (90.9%) | 130 (84.4%) | 132 (85.7%) | 160 (94.1%) | 158 (86.3%) | N/A | 322 (76.8%) |
| Hispanic | 11 (1.4%) | 8 (1.7%) | 2 (1.3%) | 1 (0.6%) | 20 (5.9%) | 11 (6.0%) | N/A | 97 (23.2%) |
| Unknown | 77 (10.0%) | 34 (7.4%) | 22 (14.3%) | 21 (13.6%) | 0 | 14 (7.7%) | N/A | 0 |

[a]Synthetic Validated = sentences used to evaluate GPT models, thus there is no demographic information for this dataset.
[b]Synthetic Demo = sentences used for bias evaluation, where demographic descriptors were inserted.
N/A = not applicable
All data presented as n (%) unless otherwise noted.

*Task definition and data labeling*

We defined our label schema and classification tasks by first carrying out interviews with subject matter experts, including social workers, resource specialists, and oncologists, to determine SDoH that are clinically relevant but not already readily available as structured data in the EHR, especially as dynamic features over time. After initial interviews, a set of exploratory pilot annotations was conducted on a subset of clinical notes and preliminary annotation guidelines were developed. The guidelines were then iteratively refined and finalized based on the pilot annotations and additional input from subject matter experts. The following SDoH categories and their attributes were selected for inclusion in the project: Employment status (employed, unemployed, underemployed, retired, disability, student), Housing issue (financial status, undomiciled, other), Transportation issue (distance, resource, other), Parental status (if the patient has a child under 18 years old), Relationship (married, partnered, widowed, divorced, single), and Social support (presence or absence of social support).

We defined two multilabel sentence-level classification tasks:

1. Any SDoH mentions: The presence of language describing an SDoH category as defined above, regardless of the attribute.
2. Adverse SDoH mentions: The presence or absence of language describing an SDoH category with an attribute that could create an additional social work or resource support need for patients:
    - **Employment status**: *unemployed, underemployed, disability*
    - **Housing issue**: *financial status, undomiciled, other*
    - **Transportation issue**: *distance, resources, other*
    - **Parental status**: *having a child under 18 years old*
    - **Relationship**: *widowed, divorced, single*
    - **Social support**: *absence of social support*

After finalizing the annotation guidelines, two annotators manually annotated the RT corpus. A total of 300/800 (37.5%) of the notes underwent dual annotation. Before adjudication, dually-annotated notes had a Krippendorf's alpha agreement of 0.86 and Cohen's Kappa of 0.86 for any SDoH mention categories. For adverse SDoH mentions, notes had a Krippendorf's alpha agreement of 0.76 and Cohen's Kappa of 0.76. Detailed agreement metrics are in **Supplemental Materials, Tables A1-2**. A single annotator then annotated the remaining radiotherapy notes, the immunotherapy dataset, and the MIMIC-III dataset. **Supplemental A** includes more details on the annotation process. **Table 2** describes the distribution of labels across the datasets and the label-level inter-annotator agreement on the radiotherapy dataset.

*Data augmentation*
We employed synthetic data generation methods to assess the impact of data augmentation for the positive class, and also to enable an exploratory evaluation of proprietary large LMs that could not be used with protected health information. In round 1, GPT-turbo-0301(ChatGPT) version of GPT3.5 via the OpenAI[34] API was prompted to generate new sentences for each SDoH category, using sentences from the annotation guidelines as references. In round 2, in order to generate more linguistic diversity, the sample synthetic sentences output from round 1 were taken as references to again generate another set of synthetic sentences. One hundred sentences per category were generated in each round. **Supplemental Material, Table A4** provides full details of prompting methods.

*Synthetic test set generation*
Iteration 1 for generating SDoH sentences involved prompting the 538 synthetic sentences to be manually validated to evaluate ChatGPT, which cannot be used with protected health information. Of these, only 480 were found to have any SDoH mention, and 289 to have an adverse SDoH mention (***Table 2***). For all synthetic data generation methods, no real patient data were used in prompt development or fine-tuning.

*Table 2. Distribution of documents and sentence labels in each dataset*

| | Number of Documents | | | | | | |
|---|---|---|---|---|---|---|---|
| | Radiotherapy | | | | | Synthetic Validated[a] | Synthetic Demo[b] |
| | Train Set | Development Set | Test Set | Immunotherapy | MIMIC-III | | |
| Documents | 481 | 160 | 159 | 200 | 200 | N/A | N/A |
| | Number of Sentences - Any SDoH Mentions | | | | | | |
| | Radiotherapy | | | | | Synthetic Validated | Synthetic Demo |
| Label | Train Set (n=29,869) | Development Set (n=10,712) | Test Set (n=10,860) | Immunotherapy (n=14,761) | MIMIC-III (n=5,328) | (n=480) | (n=419) |
| No SDoH | 28992 (97.1%) | 10429 (97.4%) | 10582 (97.4%) | 14319 (97.0%) | 4968 (93.2%) | N/A | N/A |
| Employment | 218 (0.7%) | 65 (0.6%) | 64 (0.6%) | 103 (0.7%) | 70 (1.3%) | 136 (28.3%) | 132 (31.5%) |
| Housing | 20 (0.1%) | 7 (0.1%) | 4 (0.0%) | 13 (0.1%) | 3 (0.1%) | 69 (14.4%) | 64 (15.3%) |
| Parent | 53 (0.2%) | 24 (0.2%) | 22 (0.2%) | 30 (0.2%) | 27 (0.5%) | 67 (14.0%) | 43 (10.3%) |
| Relationship | 464 (1.6%) | 153 (1.4%) | 158 (1.5%) | 241 (1.6%) | 180 (3.4%) | 152 (31.7%) | 134 (32.0%) |
| Social Support | 234 (0.8%) | 51 (0.5%) | 61 (0.6%) | 86 (0.6%) | 122 (2.3%) | 102 (21.3%) | 90 (21.5%) |
| Transportation | 41 (0.1%) | 13 (0.1%) | 6 (0.1%) | 25 (0.2%) | 3 (0.1%) | 61 (12.7%) | 58 (13.8%) |
| | Number of Sentences - Adverse SDoH Mentions | | | | | | |
| | Radiotherapy | | | | | Synthetic Validated | Synthetic Demo |
| Label | Train Set (n=29,869) | Development Set (n=10,712) | Test Set (n=10,860) | Immunotherapy (n=14,761) | MIMIC-III (n=5,328) | (n=289) | (n=253) |
| No Adverse SDoH | 29550 (98.9%) | 10615 (99.1%) | 10761 (99.1%) | 14621 (99.1%) | 5213 (97.8%) | N/A | N/A |
| Employment | 93 (0.3%) | 23 (0.2%) | 30 (0.3%) | 37 (0.3%) | 39 (0.7%) | 40 (13.8%) | 39 (15.4%) |
| Housing | 20 (0.1%) | 7 (0.1%) | 4 (0.0%) | 13 (0.1%) | 3 (0.1%) | 69 (23.9%) | 64 (25.3%) |
| Parent | 53 (0.2%) | 24 (0.2%) | 22 (0.2%) | 30 (0.2%) | 27 (0.5%) | 67 (23.2%) | 43 (17.0%) |
| Relationship | 86 (0.3%) | 27 (0.3%) | 31 (0.3%) | 30 (0.2%) | 23 (0.4%) | 68 (23.5%) | 62 (24.5%) |
| Social Support | 54 (0.2%) | 8 (0.1%) | 12 (0.1%) | 12 (0.1%) | 27 (0.5%) | 39 (13.5%) | 43 (17.0%) |
| Transportation | 41 (0.1%) | 13 (0.1%) | 6 (0.1%) | 25 (0.2%) | 3 (0.1%) | 61 (21.1%) | 58 (22.9%) |

[a]Synthetic Validated = sentences used to evaluate GPT models, thus there is no demographic information for this dataset.
[b]Synthetic Demo = sentences used for bias evaluation, where demographic descriptors were inserted.
All data presented as n (%) unless otherwise noted.
N.B. Labels sum to > 100% because some sentences had more than 1 SDoH label.
SDoH = social determinants of health; N/A = not applicable

*Model development*

The radiotherapy corpus was split into a 60%/20%/20% distribution for training, development, and testing respectively. The entire immunotherapy and MIMIC-III corpora were held-out for validation and were not used during model development.

The experimental phase of this study focused on investigating the effectiveness of different machine learning models and data settings for the classification of SDoH. We explored one multi-label BERT model as a baseline, namely bert-base-uncased[35], as well as a range of Flan-T5 models[36,37] including Flan-T5 base, large, XL, and XXL; where XL and XXL used a parameter efficient tuning method (low-rank adaptation (LoRA)[38]). Binary cross-entropy loss with logits was used for BERT and cross-entropy loss for the Flan T5 models. With Flan-T5 being a sequence-to-sequence architecture, we predicted our label-space as the target vocabulary, and post-processed it with a simple dictionary mapping (e.g., 'RELAT' → 'RELATIONSHIP'). Given the large class imbalance, non-SDoH sentences were undersampled during training. We assessed the impact of adding synthetic data on model performance.

*Ablation studies*

Ablation studies were carried out to understand the impact of manually labeled training data quantity on performance when synthetic SDoH data is included in the training dataset. First, models were trained using 10%, 25%, 40%, 50%, 70%, 75%, and 90% of manually labeled sentences; both SDOH and non-SDOH sentences were reduced at the same rate.

*Evaluation*

During training and fine-tuning, we evaluated all models using the development set and assessed their final performance on the held-out test set. For each classification task, we calculated precision/positive predictive value, recall/sensitivity, and F1 (harmonic mean of recall and precision) (***Supplemental Materials, Appendix A***). Manual error analysis was conducted on the radiotherapy dataset using the best-performing model.

*ChatGPT-family model evaluation*

To evaluate ChatGPT, the Scikit-LLM[39] multi-label zero-shot classifier and few-shot binary classifier were adapted to form a multi-label zero- and few-shot classifier (***Figure 1***). A subset of 364 unique synthetic sentences whose labels were manually validated, were used for testing. Test sentences were inserted into the prompt template, which instructs ChatGPT to act as a multi-label classifier model, and to label the sentences accordingly. Of note, because we were unable to generate high-quality synthetic non-SDoH sentences, these classifiers did not include a negative class. We evaluated the most current ChatGPT model freely available at the time of this work, GPT-turbo-0613, as well as GPT4-0314, via the OpenAI API.

```
[Context and instruction]
[Input]
[Responses]

         Prompt Example  =>   One/Few-Shot
```

> You will be provided with the following information:
> 1. An arbitrary text sample. The sample is delimited with triple backticks.
> 2. List of categories the text sample can be assigned to. The list is delimited with square brackets. The categories in the list are enclosed in the single quotes and comma separated.
> 3. Examples of text samples and their assigned categories. The examples are delimited with triple backticks. The assigned categories are enclosed in a list-like structure. These examples are to be used as training data.
>
> Perform the following tasks:
> 1. Identify to which category the provided text belongs to with the highest probability.
> 2. Assign the provided text to that category.
> 3. Provide your response in a JSON format containing a single key `label` and a value corresponding to the assigned category. Do not provide any additional information except the JSON.
>
> List of categories: {labels}
>
> Training data:
> {training_data}
>
> Text sample: ```Childcare provider offers after-school tutoring services helping child stay on track academically while parent undergoes treatment```
>
> Your JSON response:
> ==============================================================
> PARENT

*Figure 1. Example of prompt templates used in the SKLLM package for GPT-turbo-0301 (GPT3.5) and GPT4 to classify our labeled synthetic data. {labels} and {training_data} were sampled from a separate synthetic dataset, which was not human-annotated. The final label output is shown highlighted in green.*

*Language model bias evaluation*

In order to test for bias in our best-performing models and in large LMs pre-trained on general text, we used GPT4 to insert demographic descriptors into our synthetic data, as illustrated in **Figure 2**. GPT4 was supplied with our synthetically-generated test sentences, and prompted to insert demographic information into them (***Supplemental Material, Appendix A***). For example, a sentence starting with "Widower admits fears surrounding potential judgment…" might become "Hispanic widower admits fears surrounding potential judgment…". These sentences were then manually validated; 419 had any SDoH mention, and 253 had an adverse SDoH mention. THe rate of discrepant SDoH classifications with and without the injection of demographic information were compared between the best-performing fine-tuned models and ChatGPT using chi-squared tests for multi-class comparisons and 2-proportion z-tests for binary comparisons. A 2-sided $P \leq 0.05$ was considered statistically significant. Statistical analyses were carried out using the statistical Python package in scipy (Scipy.org).

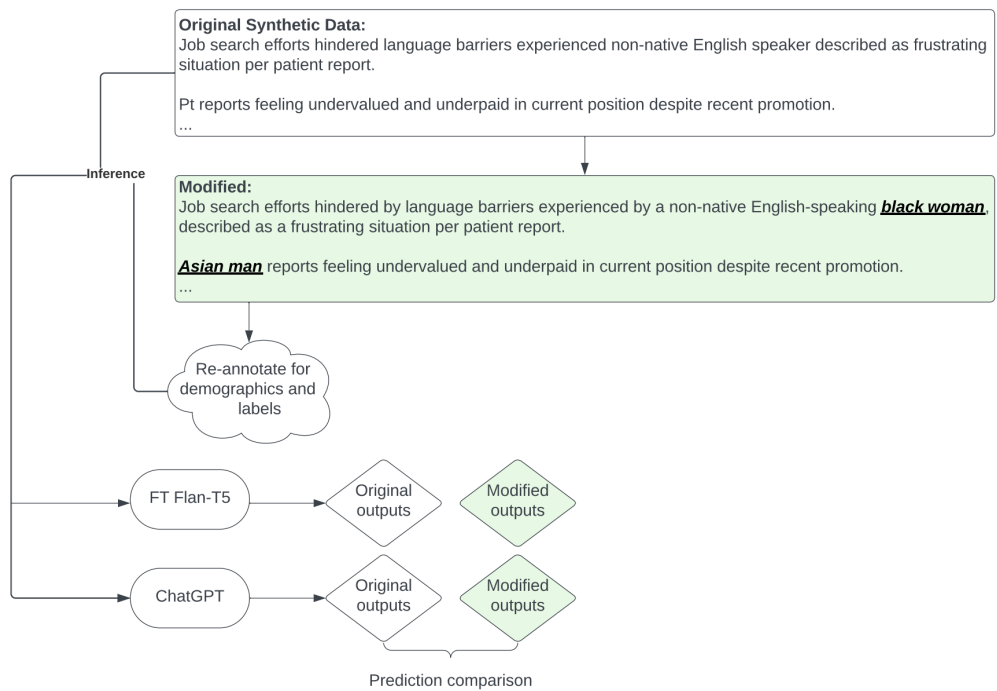

*Figure 2. Illustration of generating and comparing synthetic demographic-injected SDoH language pairs to assess how adding race/ethnicity and gender information into a sentence may impact model performance. FT = fine-tuned.*

*Comparison with structured EHR data*
To assess the completeness of SDoH documentation in structured versus unstructured EHR data, we collected Z-codes for all patients in our test set. Z-codes are SDoH-related ICD-10-CM diagnostic codes, mapped most closely with our SDoH categories present as structured data for the radiotherapy dataset (**Supplemental Materials, Table A3**). Text-extracted patient-level SDoH information was defined as the presence of one or more labels in any note. We compared these patient-level labels to structured Z-codes entered in the EHR during the same time frame.

The final annotation guidelines, analytic code, and all synthetic datasets used in this study are available at: https://github.com/AIM-Harvard/SDoH. Python version 3.9.16 (Python Software Foundation) was used to carry out this work.

## RESULTS

*Model performance*
**Table 3** shows the performance of fine-tuned models for both SDoH tasks on the radiotherapy test set. The best-performing model for any SDoH mention task was Flan-T5 XXL (3 out of 6 categories) using synthetic data (Macro-F1 0.71). The best-performing model for the adverse SDoH mention task was Flan-T5 XL without synthetic data (macro-F1 0.70). In general, the Flan-T5 models outperformed BERT, and model performance scaled with size. However, although the Flan-T5 XL and XXL models were the largest models evaluated in terms of total parameters, because LoRA was used for their fine-tuning, the fewest parameters were tuned for these models: 9.5M and 18M for Flan-TX XL and XXL, respectively compared to 110M for BERT. The negative class generally had the best performance overall, followed by Relationship and Employment. Performance varied quite a bit across the models for the other classes.

*Table 3. Model performance on the RT dataset*

| Model | Parameters (Total/Tuned) | Macro-F1 | Delta F1[a] | No SDoH | Employment | Housing | Parent | Relationship | Social Support | Transportation |
|---|---|---|---|---|---|---|---|---|---|---|
| \multicolumn{11}{c}{Any SDoH} |
| **BERT-base** | 110M/110M | | -0.01 | | | | | | | |
| Gold data only | | 0.51 | | 1.00 | 0.71 | 0.00 | 0.00 | **0.97** | 0.59 | 0.29 |
| Gold + synthetic data | | 0.49 | | 1.00 | 0.72 | 0.00 | 0.26 | 0.94 | 0.55 | 0.00 |
| **Flan-T5-base** | 250M/250M | | 0.14 | | | | | | | |
| Gold data only | | 0.36 | | 0.99 | 0.34 | 0.00 | 0.00 | 0.83 | 0.38 | 0.00 |
| Gold + synthetic data | | 0.51 | | 1.00 | 0.67 | 0.40 | 0.00 | 0.93 | 0.28 | 0.29 |
| **Flan-T5-large** | 780M/780M | | 0.18 | | | | | | | |
| Gold data only | | 0.42 | | 1.00 | 0.72 | 0.00 | 0.00 | 0.93 | 0.31 | 0.00 |
| Gold + synthetic data | | 0.61 | | 1.00 | 0.76 | 0.67 | 0.24 | 0.91 | 0.48 | 0.18 |
| **Flan-T5 XL** | 3B/9.5M | | 0.03 | | | | | | | |
| Gold data only | | 0.65 | | 0.99 | 0.71 | 0.57 | 0.55 | 0.92 | 0.50 | 0.33 |
| Gold + synthetic data | | 0.69 | | 1.00 | 0.73 | 0.55 | 0.56 | 0.94 | 0.52 | **0.53** |
| **Flan-T5 XXL** | 11B/18M | | 0.04 | | | | | | | |
| Gold data only | | **0.66** | | 1.00 | 0.76 | 0.33 | **0.65** | 0.95 | 0.51 | 0.46 |
| Gold + synthetic data | | **0.71** | | 1.00 | **0.80** | 0.67 | 0.47 | 0.93 | **0.60** | 0.47 |
| \multicolumn{11}{c}{Adverse SDoH} |
| **BERT-base** | 110M/110M | | -0.06 | | | | | | | |
| Gold data only | | 0.65 | | 1.00 | 0.68 | 0.57 | 0.43 | 0.92 | 0.45 | **0.53** |
| Gold + synthetic data | | 0.59 | | 1.00 | **0.75** | 0.57 | 0.53 | 0.82 | 0.25 | 0.22 |
| **Flan-T5-base** | 250M/250M | | 0.12 | | | | | | | |
| Gold data only | | 0.24 | | 1.00 | 0.00 | 0.00 | 0.00 | 0.43 | 0.00 | 0.29 |
| Gold + synthetic data | | 0.36 | | 1.00 | 0.31 | 0.40 | 0.00 | 0.56 | 0.00 | 0.27 |
| **Flan-T5-large** | 780M/780M | | 0.23 | | | | | | | |
| Gold data only | | 0.28 | | 0.99 | 0.46 | 0.00 | 0.00 | 0.47 | 0.00 | 0.00 |
| Gold + synthetic data | | 0.50 | | 1.00 | 0.58 | 0.55 | 0.33 | 0.66 | 0.22 | 0.18 |
| **Flan-T5 XL** | 3B/9.5M | | 0.00 | | | | | | | |
| Gold data only | | **0.70** | | 1.00 | 0.75 | 0.57 | 0.52 | **0.93** | 0.44 | 0.67 |
| Gold + synthetic data | | 0.69 | | 1.00 | 0.72 | **0.67** | 0.49 | 0.87 | **0.56** | 0.57 |
| **Flan-T5 XXL** | 11B/18M | | 0.03 | | | | | | | |
| Gold data only | | 0.64 | | 1.00 | 0.67 | 0.50 | **0.60** | 0.91 | 0.31 | 0.47 |
| Gold + synthetic data | | 0.66 | | 1.00 | 0.62 | 0.60 | 0.55 | 0.89 | 0.53 | 0.46 |

[a]Delta F1 score is the change in Macro-F1 when synthetic data is added to the fine-tuning data.
Bolded text indicates the best performance with and without synthetic data augmentation.
SDoH = social determinants of health.

For both tasks, the best-performing models with synthetic data augmentation used sentences from both rounds of GPT3.5 prompting. Synthetic data augmentation tended to lead to the largest performance improvements for classes with few instances in the training dataset and for which the model trained on gold-only data had very low performance (Housing, Parent, and Transportation).

The performance of the best-performing models for each task on the immunotherapy and MIMIC-III datasets are shown in **Table 4.** Performance was similar in the immunotherapy dataset, which represents a separate but similar patient population treated at the same hospital system. We observed a performance decrement on the MIMIC-III dataset, representing a more dissimilar patient population from a different hospital system. Performance was similar between models developed with and without synthetic data.

*Table 4. Results of the best-performing models on the out-of-domain validation datasets*

| | Any SDoH | | | | | | | | |
|---|---|---|---|---|---|---|---|---|---|
| Dataset | Macro-F1 | Delta F1[a] | No SDoH | Employment | Housing | Parent | Relationship | Social Support | Transportation |
| **Immunotherapy** | | 0.01 | | | | | | | |
| FlanXXL: Gold data only | 0.70 | | 0.99 | 0.83 | 0.56 | 0.69 | 0.93 | 0.46 | 0.46 |
| FlanXXL: Gold + synthetic data | 0.71 | | 0.99 | 0.79 | 0.56 | 0.68 | 0.91 | 0.63 | 0.40 |
| **MIMIC-III** | | -0.02 | | | | | | | |
| FlanXXL: Gold data only | 0.57 | | 0.98 | 0.65 | 0.00 | 0.63 | 0.91 | 0.32 | 0.50 |
| FlanXXL: Gold + synthetic data | 0.55 | | 0.98 | 0.69 | 0.25 | 0.44 | 0.91 | 0.33 | 0.25 |
| | Adverse SDoH | | | | | | | | |
| Dataset | Macro-F1 | Delta F1 | No SDoH | Employment | Housing | Parent | Relationship | Social Support | Transportation |
| **Immunotherapy** | | 0.03 | | | | | | | |
| FlanXL: Gold data only | 0.64 | | 1.00 | 0.70 | 0.44 | 0.62 | 0.83 | 0.42 | 0.44 |
| FlanXL: Gold + synthetic data | 0.66 | | 1.00 | 0.60 | 0.63 | 0.61 | 0.82 | 0.59 | 0.40 |
| **MIMIC-III** | | 0.00 | | | | | | | |
| FLANXL: Gold data only | 0.54 | | 0.99 | 0.55 | 0.50 | 0.37 | 0.71 | 0.37 | 0.29 |
| FLANXL: Gold + synthetic data | 0.54 | | 0.99 | 0.55 | 0.36 | 0.54 | 0.68 | 0.44 | 0.20 |

[a]Delta F1 score is the change in Macro-F1 when synthetic data is added to the fine-tuning data.
Bolded text indicates the best performance with and without synthetic data augmentation.
SDoH = social determinants of health.
Best models are determined by the performance from RT-test set

### *Ablation studies*
The ablation studies showed a consistent deterioration in model performance across all SDoH tasks and categories as the volume of real gold SDoH sentences progressively decreased, although models that included synthetic data maintained performance at higher levels throughout and were less sensitive to decreases in gold data (**Figure 3, Supplemental Material Table B1**). When synthetic data were included in training, performance was maintained until approximately 50% of gold data were removed from the train set. Conversely, without synthetic data, performance dropped after about 10-20% of the gold data were removed from the train set mimicking a true low-resource setting.

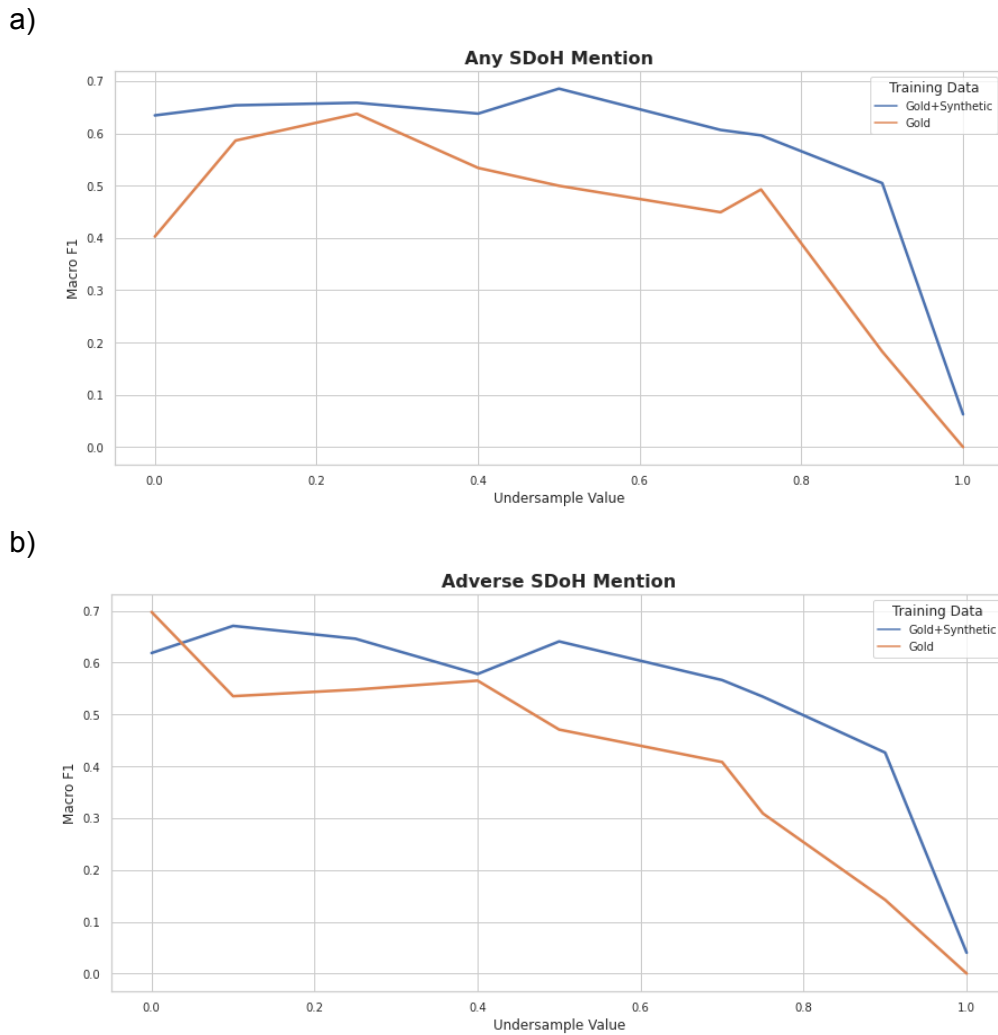

*Figure 3. Performance in Macro F1 of Flan-T5 XL models fine-tuned using gold data only (orange line) and gold and synthetic data (blue line), as gold-labeled sentences are gradually reduced by undersample value from the training dataset for the (a) any social determinant of health (SDoH) mention task and (b) adverse SDoH mention task. The full gold-labeled training set is comprised of 29,869 sentences, augmented with 1,800 synthetic SDoH sentences.*

*Error analysis*
The leading discrepancies between ground-truth and model prediction for each task are in ***Supplemental Material, Table B2***. Qualitative analysis revealed 4 distinct error patterns: Human annotator error; false positives and false negatives for Relationship and Support labels in the presence of any family mentions; incorrect labels due to information present in the note but external to the sentence and therefore not accessible to the model; and incorrectly labeling a non-adverse SDoH as an adverse SDoH.

*ChatGPT-family model performance*

When evaluating our fine-tuned Flan-T5 models on the synthetic validation dataset against GPT-turbo-0613 and GPT4-0314, our model surpassed the performance of the top-performing 10-shot learning GPT model by a margin of Macro-F1 0.05 (**Figure 4**).

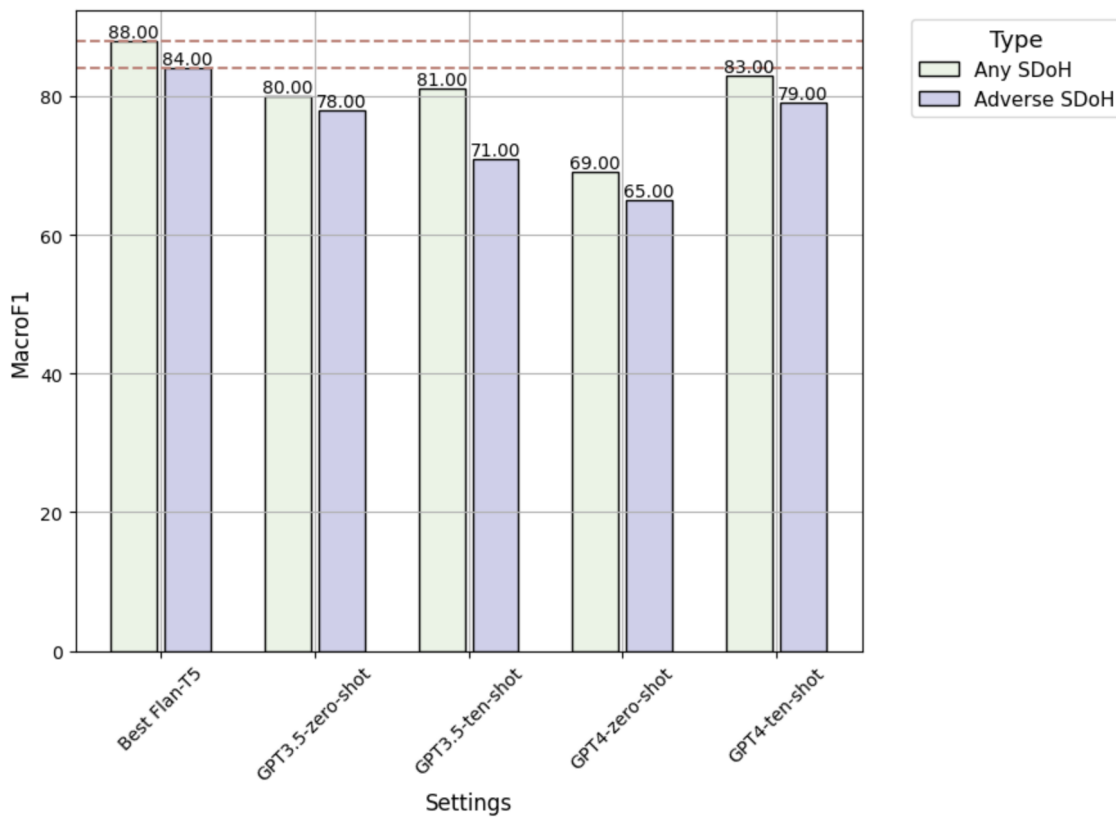

*Figure 4. This comparison shows the difference in model performance between our fine-tuned Flan-T5 models against zero- and 10-shot GPT. Macro-F1 was measured using our manually validated synthetic dataset. The GPT-turbo-0613 version of GPT3.5 and the GPT4-0314 version of GPT4 was used. The red dashed lines indicate the performance of the best-performing fine-tuned FLAN-T5 models for this task.*

*Language model bias evaluation*

Both fine-tuned Flan-T5 models and ChatGPT synthetic provided discrepant classification for sentence pairs with and without demographic information injected (**Figure 5**). However, the discrepancy rate of our fine-tuned models was nearly half that of ChatGPT: 14.3% vs. 21.5% of sentence pairs for any SDoH (P = 0.007) and 9.9% vs. 18.2% of sentence pairs for adverse SDoH (P = 0.005) for fine-tuned Flan-T5 vs. ChatGPT, respectively. ChatGPT was significantly more likely to change its classification when a female gender was injected compared to a male gender for the Any SDoH task (P = 0.01); no other within-model comparisons were statistically significant. Sentences gold-labeled as Support for both any SDoH and adverse SDoH mentions were most likely to lead to discrepant predictions for ChatGPT (56.3% (27/48)) and (21.0% (9/29)), respectively). Employment gold-labeled sentences were most-likely to lead to discrepant prediction for any SDoH mention fine-tuned model (14.4% (13/90)), and Transportation for adverse SDoH mention fine-tuned model (12.2% (6/49)).

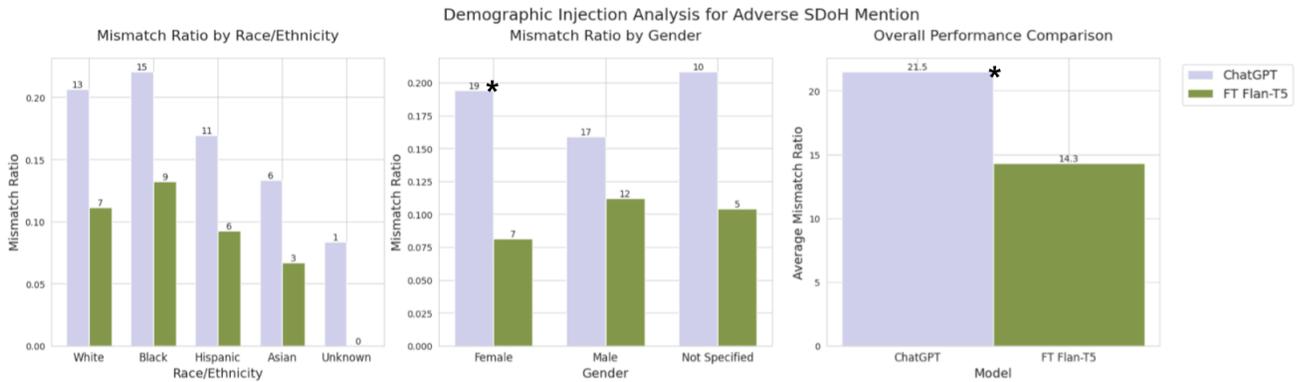

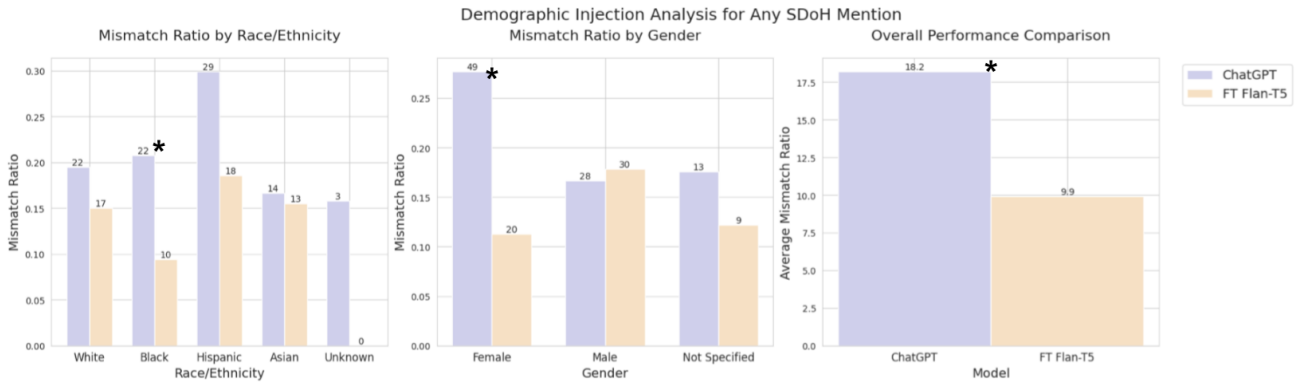

*Figure 5. The proportion of synthetic sentence pairs with and without demographics injected that led to a classification mismatch, meaning that the model predicted a different SDoH label for each sentence in the pair. Overall, Results are shown across race/ethnicity and gender for (a) adverse SDoH mention task and (b) any SDoH mention task. Asterisks indicate statistical significance (P ≤ 0.05).*

*Comparison with structured EHR data*
Our best-performing models for any SDoH mention correctly identified 95.7% (89/93) patients with at least one SDoH mention, and 93.8% (45/48) patients with at least one adverse SDoH mention (***Supplemental Material, Tables B3-4***). SDoH entered as structured Z-code in the EHR during the same timespan identified 2.0% (1/48) with at least one adverse SDoH mention (all mapped Z-codes were adverse) (***Supplemental Material, Table B5***). ***Supplemental Material, Figures B1-2*** shows that patient-level performance when using model predictions out-performed Z-codes by a factor of at least 3 for every label for each task (Macro-F1 0.78 vs. 0.17 for any SDoH mention and 0.71 vs. 0.17 for adverse SDoH mention).

## DISCUSSION

We developed multilabel classifiers to identify the presence of 6 different SDoH documented in clinical notes, demonstrating the potential of large LMs to improve the collection of real-world data on SDoH and support appropriate allocation of resource support to patients who need it most. We identified a substantial performance gap between a more traditional BERT classifier and larger Flan-T5 XL and XXL models. Our fine-tuned models out-performed ChatGPT-family models with zero- and few-shot learning, and were less sensitive to the injection of demographic descriptors. Compared to diagnostic codes entered as structured data, text-extracted data identified 91.8% more patients with an adverse SDoH. We also contribute **new annotation guidelines** as well as **synthetic SDoH datasets** to the research community.

All of our models performed well at identifying sentences that do not contain SDoH mentions (F1 ≥ 0.99 for all). For any SDoH mentions, performance was worst for parental status and transportation issues. For adverse SDoH mentions, performance was worst for parental status and social support. These findings are unsurprising given the marked class imbalance for all SDoH labels—Only **3%** of sentences in our training set contained any SDoH mention. Given this imbalance, our models' ability to identify sentences that contain SDoH language is impressive. In addition, these SDoH descriptions are semantically and linguistically complex. In particular, sentences describing social support are highly variable given the variety of ways individuals can receive support from their social systems during care. Interestingly, our best-performing models demonstrated strong performance in classifying housing issues (macro-F1 0.67), which was our scarcest label with only 20 instances in the training dataset. This speaks to the potential of large LMs in improved real-world data collection for very sparsely documented information, which is the most likely to be missed via manual review.

The recent advancements in large LMs have opened a pathway for synthetic text generation that may improve model performance via data augmentation, and enable experiments that better protect patient privacy.[40] This is an emerging area of research that falls within a larger body of work on synthetic patient data across a range of data types and end-uses.[41,42] **Our study is among the first to evaluate the role of contemporary generative large LMs for synthetic clinical text to help unlock the value of unstructured data within the EHR.** We were particularly interested in synthetic clinical data as a means to address the aforementioned scarcity of SDoH documentation, and our findings may provide generalizable insights for the common clinical NLP challenges of class imbalance—Many clinically important data are difficult to identify among the huge amounts of text in a patient's EHR. We found variable benefits of synthetic data augmentation across model architecture and size; the strategy was most beneficial for the smaller Flan-T5 models and for the rarest classes where performance was dismal using gold data alone. Importantly, the ablation studies demonstrated that only approximately half of the gold-labeled dataset was needed to maintain performance when synthetic data was included in training, although synthetic data alone did not produce high-quality models. Of note, we aimed to understand whether synthetic data for augmentation could be automatically generated using ChatGPT-family models without additional human annotation, and so it is possible that manual gold-labeling could further enhance the value of these data. However, this would decrease the value of synthetic data in terms of reducing annotation effort.

Our novel approach to generating synthetic clinical sentences also enabled us to explore the potential for ChatGPT-family models, GPT3.5 and GPT4, for supporting the collection of SDoH information from the EHR. We found that fine-tuning LMs that are orders of magnitude smaller than ChatGPT-family models, even with our relatively small dataset, outperformed zero-shot and few-shot learning with ChatGPT-family models, consistent with prior work evaluating large LMs for clinical uses.[43,44] Nevertheless, GPT3.5 in particular showed promising performance for a model that was not explicitly trained for clinical tasks.

It is well-documented that LMs learn the biases, prejudices, and racism present in the language they are trained on.[45–48] Thus, it is essential to evaluate how LMs could propagate existing biases, which in clinical settings could amplify the health disparities epidemic.[1–3] We were especially concerned that SDoH-containing language may be especially prone to eliciting these biases. Both our fine-tuned models and ChatGPT altered their SDoH classification predictions when demographics and gender descriptors were injected into sentences, although the fine-tuned models were significantly more robust that ChatGPT. Although not significantly different, it is worth noting that for both the fine-tuned models and ChatGPT, **Hispanic and Black descriptors were most likely to change the classification for any SDoH and adverse SDoH mentions, respectively**. This lack of significance may be due to the small numbers in this evaluation, and future work is critically needed to further evaluate bias in clinical LMs. We have made our paired demographic-injected sentences openly available for future efforts on LM bias evaluation.

SDoH are notoriously under-documented in existing EHR structured data.[10–12,49] Our findings that **text-extracted SDoH information was better able to identify patients with adverse SDoH than relevant billing codes** are in agreement with prior work showing under-utilization of Z-codes[10,11]. Most EMR systems have other ways to enter SDoH information as structured data which may have more complete documentation, however these did not exist for most of our target SDoH. Lyberger et al. evaluated other EHR sources of structured SDoH data, and similarly found that NLP methods are a complementary source SDoH information extraction, and were able to identify 10-30% of patients with tobacco, alcohol, and homelessness risk factors documented only in unstructured text[22].

There have been several prior studies developing NLP methods to extract SDoH from the EHR.[13–21,50] The most common SDoH targeted in prior efforts include smoking history, substance use, alcohol use, and homelessness.[23] In addition, many prior efforts focus only on text in the Social History section of notes. In a recent shared task on alcohol, drug, tobacco, employment, and living situation event extraction from Social History sections, pre-trained LMs similarly provided best performance.[51] In addition to our technical innovations, our work adds to prior efforts by investigating SDoH that are less commonly targeted for extraction but nonetheless have been shown to impact healthcare.[52–60] We also developed methods that can mine information from full clinic notes, not only Social History sections—a fundamentally more challenging task with a much larger class imbalance. Clinically-impactful SDoH information is often scattered throughout other note sections and many note types, such as many inpatient progress notes and notes written by nurses and social workers, do not consistently contain Social History sections.

Our study has limitations. First, our training and validation datasets come from a predominantly white population treated at hospitals in Boston, Massachusetts in the United States of America. This limits the generalizability of our findings. We could not exhaustively assess the many methods to generate synthetic data from ChatGPT. Instead, we chose to investigate prompting methods that could be easily reproduced by others and did not require extensive task-specific optimization, as this is likely not feasible for the many clinical NLP tasks that one may wish to generate synthetic data on. Incorporating real clinical examples in the prompt would likely improve the quality of the synthetic data, and is an area of future research when large generative LMs become more widely available for use with protected health information and within the resource constraints of academic researchers and healthcare systems. Because we could not evaluate ChatGPT-family models using protected health information, our evaluations are limited to manually-verified synthetic sentences. Thus, our reported performance may not completely reflect true performance on real clinical text. Because the synthetic sentences were generated using ChatGPT itself, and ChatGPT presumably has not been trained on clinical text, we hypothesize that if anything, performance would be worse on real clinical data.

**CONCLUSIONS**

Our findings highlight the potential of large LMs to improve real-world data collection and identification of SDoH from the EHR. In addition, synthetic clinical text generated by large LMs may enable better identification of rare events documented in the EHR, although more work is needed to optimize generation methods. Our fine-tuned models out-performed and were less prone to bias than ChatGPT-family models, despite being orders of magnitude smaller. In the future, these models could improve our understanding of drivers of health disparities by improving real-world evidence, and could directly support patient care by flagging patients who may benefit most from proactive resource and social work referral.

## Appendix A: Supplemental Methods

*Note Selection*

**All notes:** We excluded notes with fewer than 150 tokens across all provider types. For notes written by physicians and nurses, we excluded any notes containing a single section longer than 500 tokens to avoid irrelevant or redundant information. Additionally, we only included physician notes that contained at least one of the following sections: Assessment and Plan, Social History, and History/Subjective.

**Radiotherapy (RT) Dataset:** We established a date range, considering notes within a window of 30 days before the first treatment and 90 days after the last treatment. We included notes from various healthcare professionals such as Physicians, Physician Assistants, Nurse Practitioners, Registered Nurses, and Social Workers. Additionally, in the fifth round of annotation, we specifically excluded notes from patients with zero social work notes. This decision ensured that we focused on individuals who had received social work intervention or had pertinent social context documented in their notes.

**Immunotherapy Dataset:** Only notes written by physicians, physician assistants, nurse practitioners, registered nurses, and social workers were included for analysis. We ensured that there was no patient overlap between radiation and immunotherapy notes. We also specifically selected notes from patients with at least one social work note. To further refine the selection, we considered notes with a note date one month before or after the patient's first social work note or after it.

**MIMIC-III dataset:** Only notes written by physicians, social workers, and nurses were included for analysis from the MIMIC III corpus. We focused on patients who had at least one social work note, without any specific date range criteria.

*Annotation Process*

In total, ten thousand one-hundred clinical notes were annotated line-by-line using the annotation software Multi-document Annotation Environment (MAE)[1]. The first 300 notes were jointly annotated by two data scientists across 4 rounds. After each round, the data scientists and an oncologist performed discussion-based adjudication. The data scientists then each independently annotated 449 and 351 notes, including the validation sets, for a total of 1,100 annotated notes.

1. Multi-document annotation environment. *MAE* http://keighrim.github.io/mae-annotation/.

**Table A1.** Inter-annotator agreement for granular SDoH levels

| Granular SDoH Label | Class-wise Krippendorff's (α) |
|---|---|
| TRANSPORTATION_distance | 0.00 |
| TRANSPORTATION_resource | 0.67 |
| TRANSPORTATION_other | 0.00 |
| HOUSING_undomiciled | 0.00 |
| HOUSING_poor | 0.29 |
| HOUSING_other | 0.36 |
| RELATIONSHIP_married | 0.95 |
| RELATIONSHIP_partnered | 0.93 |
| RELATIONSHIP_divorced | 0.86 |
| RELATIONSHIP_widowed | 1.00 |
| RELATIONSHIP_single | 0.84 |
| PARENT | 0.81 |
| EMPLOYMENT_employed | 0.80 |
| EMPLOYMENT_underemployed | 0.40 |
| EMPLOYMENT_unemployed | 0.74 |
| EMPLOYMENT_disability | 0.90 |
| EMPLOYMENT_retired | 0.90 |
| EMPLOYMENT_student | 1.00 |
| SUPPORT_plus | 0.78 |
| SUPPORT_minus | 0.74 |

SDoH = social determinant of health.

**Table A2.** Inter-annotator agreement for higher-level SDoH mention labels

| Any SDoH Mention Label | Class-wise Krippendorff's (α) |
|---|---|
| SUPPORT | 0.77 |
| EMPLOYMENT | 0.89 |
| HOUSING | 0.71 |
| TRANSPORTATION | 0.64 |
| PARENT | 0.81 |
| RELATIONSHIP | 0.95 |

SDoH = social determinant of health.

**Table A3.** Z-Code to SDoH label mappings

| Z-Code | SDoH label mapping |
|---|---|
| Z55: Education and literacy | EDUCATION_none |
| Z56: Employment and unemployment | EMPLOYMENT_unemployed, EMPLOYMENT_underemployed, EMPLOYMENT_employed (with adverse work environment) |
| Z59: Housing and economic circumstances | HOUSING, SUPPORT_minus, EMPLOYMENT_un(der)employed |
| Z60: Social environment | SUPPORT_minus |
| Z62: Upbringing | PARENT, SUPPORT_minus |
| Z63: Other problems related to primary support group, including family circumstances | SUPPORT_minus |
| Z75: Problems related to medical facilities and other health care | HOUSING, TRANSPORTATION |

SDoH = social determinant of health.

**Table A4.** Prompts used to generate synthetic SDoH sentences using GPT3.5

| Output Sentence Label[a] | Prompt |
|---|---|
| Housing-Adverse | {"role": "system", "content": "You are a physician."},<br>{"role": "user", "content": "Examples of housing issues for patients: 1. Pt came from Assisted Living Corp. and complained about rent increase.\n2. "Pt came from Assisted Living Corp. and complained about rent increase.\n3. He says he is worried about making his mortgage payments.\n4. Pt is staying with a friend and does not have a mailing address.\n5. Pt currently staying at Barbara McInnis shelter.\n5. Pt is staying at the Motel for the time being, while on the waitlist for the Hope Lodge."},<br>{"role": "assistant", "content": "Ok I will remember that."},<br>{"role": "user", "content": "Imagine you are a physician. Please give me 100 sentences from your clinic notes about various patient's housing issues similar to the examples."} |
| Transportation-Adverse | {"role": "system", "content": "You are a physician."},<br>{"role": "user", "content": "Examples of transportation issues for patients: 1. Pt lives 30mi away from hospital and complains about needing to transfer three times each way.\n2. Pt missed appointment because her sister couldn't drive her today.\n3. Pt is worried about making appointments because the metro is under construction this month.\n4. Pt is worried about the two hour drive.\n5. She is having trouble lying flat for treatment, she thinks it is because her back hurts after the two hour car ride into clinic.\n6. Pt felt that coming to Los Angeles was hard for them and asked to be referred to Santa Cruz.\n7. He is having trouble getting to and from the hospital."},<br>{"role": "assistant", "content": "Ok I will remember that."},<br>{"role": "user", "content": "Imagine you are a physician. Please give me 100 sentences from your clinic notes about various patient's transportation issues similar to the examples."} |
| Relationship-Adverse | {"role": "system", "content": "You are a physician."},<br>{"role": "user", "content": "Examples of divorced, widowed, single, separated issues for patients: 1. Pt is meeting ex-wife at appointment.\n2. Pt is married but separated.\n3. Pt spouse passed away in October of last year.\n4. Pt is single.\n5. Pt arrived with his girlfriend, and his ex-wife will attend with him at next week's session.\n6. Pt has 3 kids from former marriage"},<br>{"role": "assistant", "content": "Ok I will remember that."},<br>{"role": "user", "content": "Imagine you are a physician. Please give me 100 sentences from your clinic notes about various patients being divorced, widowed, single, or separated issues similar to the examples."} |
| Relationship-Not adverse | {"role": "system", "content": "You are a physician."},<br>{"role": "user", "content": "Examples of married/partnered sentences for patients: 1. Pt and her husband came into my office today.\n2. Pt and her fiancée came into my office today.\n3. He is here with his boyfriend.\n4. He is married to Sheila."},<br>{"role": "assistant", "content": "Ok I will remember that."}, |
| | {"role": "user", "content": "Imagine you are a physician. Please give me 100 sentences from your clinic notes about various patients being married / partnered similar to the examples."} |
| Parent-Adverse | {"role": "system", "content": "You are a physician."}, {"role": "user", "content": "Examples of parental status for patients: 1. Pt has 2 children ages 9 and 13.\n 2. Pt has 2 teenage children.\n3. Pt was seen today with his daughter Angela, 3 y/o for a routine checkup."}, {"role": "assistant", "content": "Ok I will remember that."}, {"role": "user", "content": "Imagine you are a physician. Please give me 100 sentences from your clinic notes about various patients being a parent to minors similar to the examples."} |
| Employment-Adverse | {"role": "system", "content": "You are a physician."}, {"role": "user", "content": "Examples of employment issues for patients: 1. Pt works part-time at Jim's Fish and is struggling to pay rent.\n2. Pt has been living off of unemployment for the past 2 months.\n3. Used to be a car mechanic, but he has been on disability for the past 2 years since his diagnosis.\n4. He is currently on disability and is also occasionally working as an Uber driver to help cover the bills."}, {"role": "assistant", "content": "Ok I will remember that."}, {"role": "user", "content": "Imagine you are a physician. Please give me 100 sentences from your clinic notes about various patient's employment issues similar to the examples."} |
| Employment-Not adverse | {"role": "system", "content": "You are a physician."}, {"role": "user", "content": "Examples of employment sentences for patients: 1. Pt works as an electrician in Rockland.\n2. Pt is a 75yr old retiree.\n3. Pt is attending Cool University full time.\n4. Pt is a semi-retired marketing consultant."}, {"role": "assistant", "content": "Ok I will remember that."}, {"role": "user", "content": "Imagine you are a physician. Please give me 100 sentences from your clinic notes about various patient's employment similar to the examples."} |
| Social support-Adverse | {"role": "system", "content": "You are a physician."}, {"role": "user", "content": "Examples of social support issues for patients: 1. Pt lives alone.\n2. Pt is struggling to find someone to watch his cat on the days he has to come for treatment."}, {"role": "assistant", "content": "Ok I will remember that."}, {"role": "user", "content": "Imagine you are a physician. Please give me 100 sentences from your clinic notes about various patient's lack of social support similar to the examples."} |
| Social support-Not adverse | {"role": "system", "content": "You are a physician."}, {"role": "user", "content": "Examples of social support sentences for patients: 1. Here today is Pt, her daughter, and supportive wife.\n2. Pt is living with his parents during treatment, while his neighbors watch his cat.\n3. Pt had to borrow money from her friend to catch the bus today.\n4. Pt is currently living with nephew while receiving treatment."}, {"role": "assistant", "content": "Ok I will remember that."}, |

| | {"role": "user", "content": "Imagine you are a physician. Please give me 100 sentences from your clinic notes about various patient's social support similar to the examples."} |
|---|---|

[a]Output sentence label is the label assigned to all synthetic sentences generated from the prompt.

**Few-shot prompt used for the GPT classifier experiments:**
Prompts were developed for each exemplar in the development/exemplar set using the following template. The final string was input into GPT-3.5 and GPT-4 for classification.

```
"Sample input: [TEXT]

Sample target: [LABELS]"
```

`[TEXT]` was the exemplar from the development/exemplar set.

`[LABELS]` was a comma-separated list of the labels for that exemplar, e.g. `PARENT,RELATIONSHIP`.

**Prompt used to generate demographic alterations for synthetic SDoH sentences:**
Demographic-altered synthetic sentences were generated for each synthetic original sentence using the following template (in a batch of 10 to ensure demographic variations):

```
"role": "user", "content": "[ORIGINAL SENTENCE]\n swap the sentences
patients above to one of the race/ethnicity [asian, black, white,
hispanic] and gender, and put the modified race and gender in bracket at
the beginning like this \n Owner operator food truck selling gourmet
grilled cheese sandwiches around town => \n [Asian female] Asian woman
owner operator of a food truck selling gourmet grilled cheese sandwiches
around town"
```

`[ORIGINAL SENTENCE]` was a sentence from a selected subset of our GPT-3.5-generated synthetic data

*Performance Metric Equations*

Precision = TP/(TP+FP)

Recall = TP/(TP+FN)

F1 = (2*Precision*Recall)/(Precision+Recall)

TP = true positives, FP = false positives, FN = false negatives

# Appendix B: Supplemental Results

**Table B1.** Ablation studies of removing gold-labeled sentences in training

| | Any SDoH | | | | | | | |
|---|---|---|---|---|---|---|---|---|
| Percent undersampled[a] | Macro-F1 | No SDoH | Employment | Housing | Parent | Relationship | Social Support | Transportation |
| 10% | | | | | | | | |
| Gold data only | 0.586 | 0.996 | 0.745 | 0.333 | 0.000 | 0.947 | 0.484 | 0.600 |
| Gold + synthetic data | 0.654 | 0.996 | 0.809 | 0.600 | 0.452 | 0.919 | 0.566 | 0.235 |
| 25% | | | | | | | | |
| Gold data only | 0.638 | 0.997 | 0.750 | 0.571 | 0.160 | 0.969 | 0.516 | 0.500 |
| Gold + synthetic data | **0.659** | 0.996 | 0.822 | 0.545 | 0.452 | 0.938 | 0.571 | 0.286 |
| 40% | | | | | | | | |
| Gold data only | 0.534 | 0.995 | 0.719 | 0.000 | 0.087 | 0.932 | 0.604 | 0.400 |
| Gold + synthetic data | 0.638 | 0.996 | 0.810 | 0.667 | 0.400 | 0.956 | 0.455 | 0.182 |
| 50% | | | | | | | | |
| Gold data only | 0.500 | 0.995 | 0.683 | 0.000 | 0.083 | 0.957 | 0.559 | 0.222 |
| Gold + synthetic data | 0.686 | 0.996 | 0.824 | 0.750 | 0.465 | 0.935 | 0.629 | 0.200 |
| 70% | | | | | | | | |
| Gold data only | 0.449 | 0.995 | 0.657 | 0.000 | 0.167 | 0.900 | 0.425 | 0.000 |
| Gold + synthetic data | 0.607 | 0.994 | 0.720 | 0.545 | 0.400 | 0.825 | 0.429 | 0.333 |
| 75% | | | | | | | | |
| Gold data only | 0.493 | 0.996 | 0.746 | 0.000 | 0.240 | 0.933 | 0.533 | 0.000 |
| Gold + synthetic data | 0.596 | 0.995 | 0.782 | 0.500 | 0.074 | 0.886 | 0.537 | 0.400 |
| 90% | | | | | | | | |
| Gold data only | 0.183 | 0.988 | 0.000 | 0.000 | 0.000 | 0.259 | 0.031 | 0.000 |
| Gold + synthetic data | 0.505 | 0.994 | 0.733 | 0.400 | 0.143 | 0.849 | 0.108 | 0.308 |
| 100% | | | | | | | | |
| Gold data only | 0.00 | 0.00 | 0.00 | 0.00 | 0.00 | 0.00 | 0.00 | 0.00 |
| Gold + synthetic data | 0.063 | 0.000 | 0.047 | 0.017 | 0.228 | 0.036 | 0.095 | 0.015 |

| | | Adverse SDoH | | | | | | |
|---|---|---|---|---|---|---|---|---|
| Percent undersampled[a] | Macro-F1 | No SDoH | Employment | Housing | Parent | Relationship | Social Support | Transportation |
| 10% | | | | | | | | |
| Gold data only | 0.535 | 0.997 | 0.746 | 0.400 | 0.083 | 0.933 | 0.143 | 0.444 |
| Gold + synthetic data | **0.671** | 0.997 | 0.719 | 0.750 | 0.474 | 0.921 | 0.375 | 0.462 |
| 25% | | | | | | | | |
| Gold data only | 0.548 | 0.997 | 0.643 | 0.571 | 0.000 | 0.881 | 0.143 | 0.600 |
| Gold + synthetic data | 0.646 | 0.997 | 0.716 | 0.667 | 0.378 | 0.879 | 0.353 | 0.533 |
| 40% | | | | | | | | |
| Gold data only | 0.565 | 0.998 | 0.730 | 0.000 | 0.410 | 0.933 | 0.286 | 0.600 |
| Gold + synthetic data | 0.578 | 0.997 | 0.635 | 0.545 | 0.400 | 0.903 | 0.300 | 0.267 |
| 50% | | | | | | | | |
| Gold data only | 0.471 | 0.997 | 0.688 | 0.571 | 0.000 | 0.755 | 0.000 | 0.286 |
| Gold + synthetic data | 0.641 | 0.997 | 0.690 | 0.667 | 0.313 | 0.921 | 0.400 | 0.500 |
| 70% | | | | | | | | |
| Gold data only | 0.408 | 0.997 | 0.500 | 0.000 | 0.160 | 0.808 | 0.143 | 0.250 |
| Gold + synthetic data | 0.566 | 0.997 | 0.646 | 0.600 | 0.389 | 0.929 | 0.154 | 0.250 |
| 75% | | | | | | | | |
| Gold data only | 0.309 | 0.996 | 0.256 | 0.000 | 0.000 | 0.410 | 0.000 | 0.500 |
| Gold + synthetic data | 0.534 | 0.996 | 0.480 | 0.545 | 0.294 | 0.866 | 0.273 | 0.286 |
| 90% | | | | | | | | |
| Gold data only | 0.142 | 0.995 | 0.000 | 0.000 | 0.000 | 0.000 | 0.000 | 0.000 |
| Gold + synthetic data | 0.426 | 0.996 | 0.311 | 0.462 | 0.160 | 0.747 | 0.143 | 0.167 |
| 100% | | | | | | | | |
| Gold data only | 0.00 | 0.00 | 0.00 | 0.00 | 0.00 | 0.00 | 0.00 | 0.00 |
| Gold + synthetic data | 0.040 | 0.000 | 0.009 | 0.007 | 0.227 | 0.011 | 0.015 | 0.011 |

Settings: FlanXL, guideline synthetic data OR no synthetic data.
[a]% undersampled means percent taken away (e.g., 25% undersampled = 25% of positive gold-labeled instances and 25% of negative gold-labeled instances removed).
SDoH = social determinants of health.

**Table B2.** Most common discrepancies between ground-truth and best-performing model prediction for each task

| Task | Ground Truth | Model Prediction | Count |
|---|---|---|---|
| **Any SDoH Mention** | No SDoH | Support | 24 |
| | No SDoH | Employment | 16 |
| | Support | No SDoH | 10 |
| **Adverse SDoH Mention** | No SDoH | Employment | 12 |
| | Parent | No SDoH | 10 |
| | Employment | No SDoH | 6 |

SDoH = social determinants of health.

**Table B3.** Confusion matrix for any SDoH mention gold label versus best-performing model prediction

| | | Gold Label | |
|---|---|---|---|
| | | Positive | Negative |
| **Any SDoH Model Prediction** | Positive | 89 | 3 |
| | Negative | 4 | 58 |

SDoH = social determinants of health.

**Table B4.** Confusion matrix for adverse SDoH mention gold label versus best-performing model prediction

| | | Gold Label | |
|---|---|---|---|
| | | Positive | Negative |
| **Adverse SDoH Model Prediction** | Positive | 45 | 13 |
| | Negative | 3 | 93 |

SDoH = social determinants of health.

**Table B5.** Confusion matrix for adverse SDoH gold label versus mapped Z-codes

| | | Gold Label | |
|---|---|---|---|
| | | Positive | Negative |
| **Mapped Z-codes** | Positive | 1 | 5 |
| | Negative | 47 | 101 |

SDoH = social determinants of health.

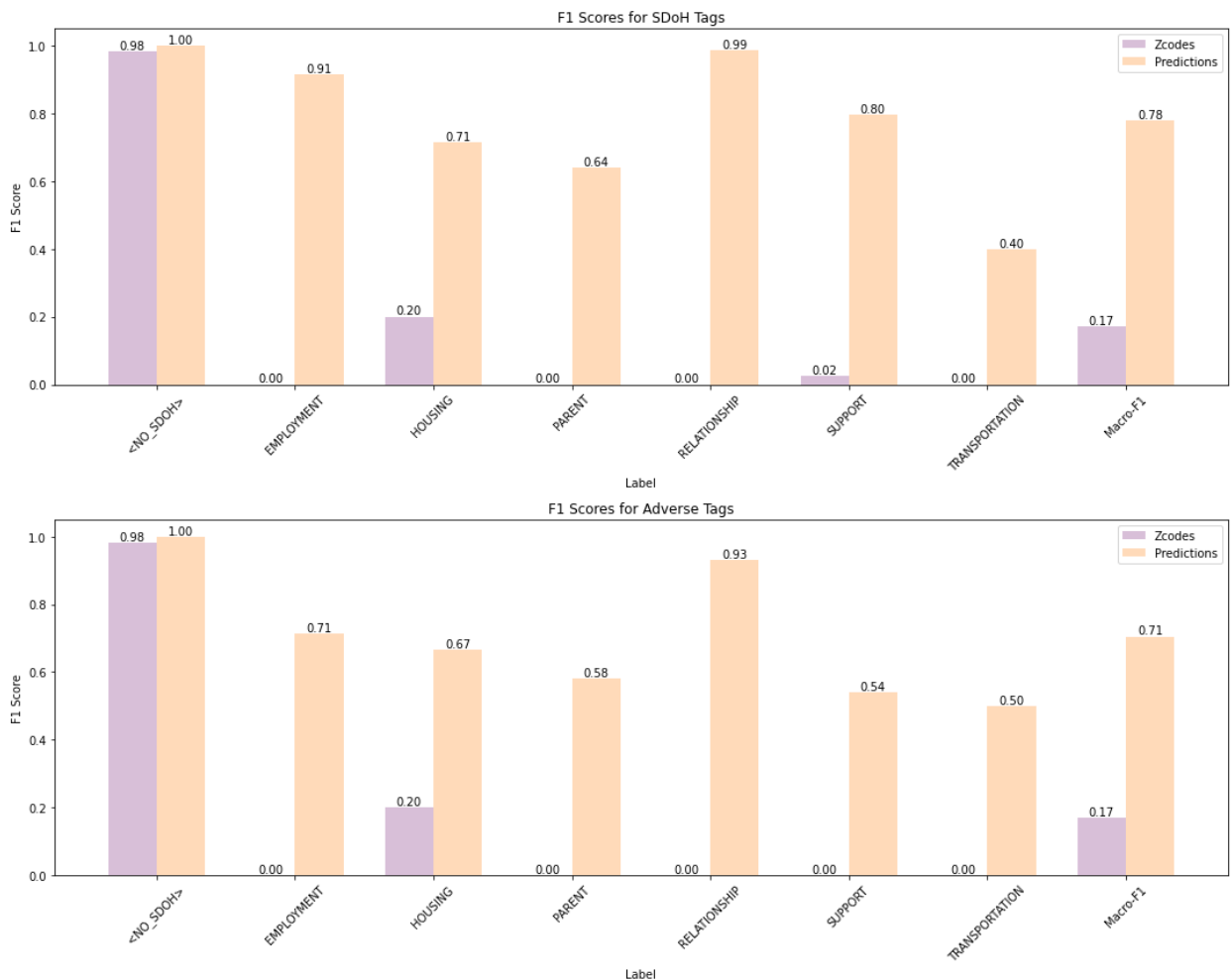

**Figure B1.** Class-wise and Macro-F1 scores of our best-performing model against mapped Z-Codes at the patient level (on test set and dev set).

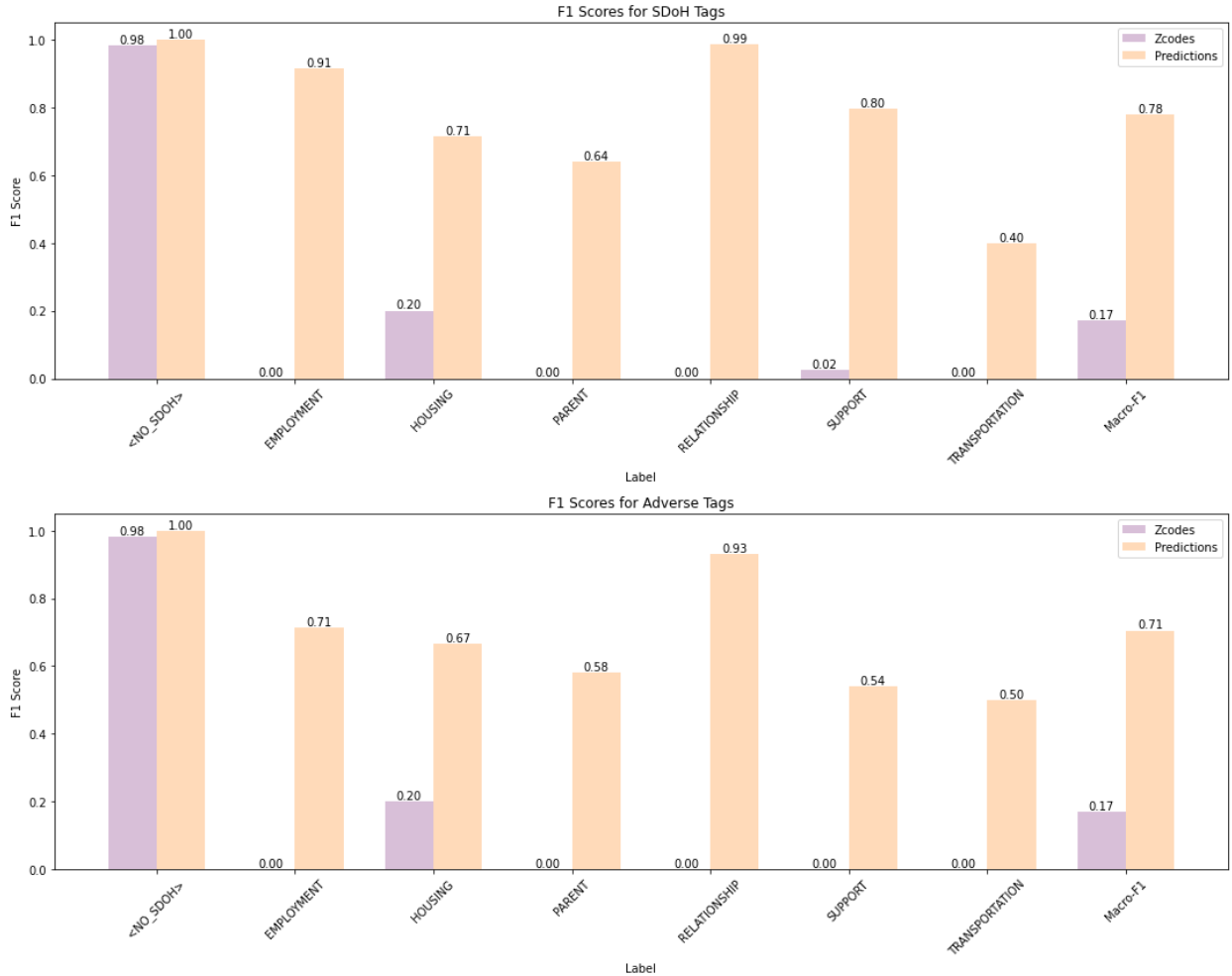

**Figure B2.** Class-wise and Macro-F1 scores of our best-performing model against mapped Z-Codes at the patient level (on test set only).